\documentclass{article}
\usepackage{spconf,amsmath,graphicx}
\usepackage{multirow}
\usepackage{enumerate}
\usepackage{multicol}

\title{Hiding Images into Images with Real-world Robustness}
\name{\normalsize{Qichao Ying$^{\star}$, Hang Zhou$^{\dagger}$, Xianhan Zeng$^{\star}$, Haisheng Xu$^{\ddagger}$, Zhenxing Qian$^{\star}$ and Xinpeng Zhang$^{\star}$\thanks{This work is supported by National Natural Science Foundation of China under Grant U20B2051, U1936214. Corresponding author: Hang Zhou and Zhenxing Qian (zhouhang2991@gmail.com, zxqian@fudan.edu.cn)}}
\address{$^{\star}$ Fudan University, $^{\dagger}$ Simon Fraser University, $^{\ddagger}$ NVIDIA}}
\usepackage{lipsum}
\usepackage{etoolbox}
\begin{document}
\ninept
%
\maketitle
\begin{abstract}
The existing image embedding networks are basically vulnerable to malicious attacks such as JPEG compression and noise adding, not applicable for real-world copyright protection tasks. To solve this problem, we introduce a generative deep network based method for hiding images into images while assuring high-quality extraction from the destructive synthesized images. An embedding network is sequentially concatenated with an attack layer, a decoupling network and an image extraction network. The addition of decoupling network learns to extract the embedded watermark from the attacked image. We also pinpoint the weaknesses of the adversarial training for robustness in previous works and build our improved real-world attack simulator. Experimental results demonstrate the superiority of the proposed method against typical digital attacks by a large margin, as well as the performance boost of the recovered images with the aid of progressive recovery strategy. Besides, we are the first to robustly hide three secret images.
\end{abstract}
\begin{keywords}
image hiding, robustness, watermarking, covert communication, generative adversarial networks
\end{keywords}

\section{Introduction}
The development of data hiding is of increasing importance to applications such as law enforcement and military communities. Besides, information hiding can also be used in copyright protection, as one embeds information into the digital contents which can declare the ownership. In the past decades, many researches have been developed for information hiding, which can be categorized into steganography \cite{fang2018screen,agarwal2019survey,asikuzzaman2017overview} and watermarking \cite{tao2018towards,sedighi2015content,chen2018defining}. 

Steganography aims at embedding large-capacity data while maintaining the invariance of the statistical characteristics of the stego image by resisting steganalysis, a two-class classification task. Recently, with the rapid development of deep learning technologies, some researchers attempt to embed larger payloads into the cover images using deep neural networks, namely, hiding an image into another image (HI3).
The technology is first proposed by Baluja \cite{baluja2017hiding} where one or multiple images can be efficiently hidden into another image called cover image using deep networks. The cover image is first processed by the preparation network to generate a high-level representation. Then, an embedding network and a reveal network are proposed for image hiding and extraction, respectively. The work soon receives extensive attentions. Some improved schemes are proposed to provide better embedding performances \cite{baluja2019hiding,duan2019reversible,rahim2018end}. 
Although previous HI3 schemes \cite{baluja2017hiding,baluja2019hiding,duan2019reversible,rahim2018end} have high capacities, they are not designed for robust data hiding. Once the marked image is tampered or compressed, the hidden information cannot be extracted correctly. 

Watermarking focuses on robustness against possible digital attacks, such as image compression, noise adding, filtering, etc. The technology has been widely used in copyright protection and content authentication of images in multimedia. Recently, many novel watermarking schemes with deep networks are proposed. Zhu et al. \cite{zhu2018hidden} proposes an end-to-end scheme for robust information hiding. The scheme is based on fully convolutional networks, which hides a fixed amount of data into a cover image. The achieved robustness is also much higher compared to traditional methods. Afterwards, robust data hiding using deep network is mainly developed for the application of digital watermarking \cite{kandi2017exploring,mun2019finding}. These schemes are both robust against some typical attacks such as JPEG compression, scaling and some degree of rotation. However, the payloads are significantly lower than the typical HI3 schemes. For example, \cite{zhu2018hidden} can only accommodate around 100 bits, while \cite{mun2019finding} only hides binary watermarks. Previously, robust watermarking has been introduced to aid many other computer vision tasks, e.g., preventing images from being inpainted \cite{khachaturov2021markpainting} or reconstructed by super-resolution \cite{yin2018deep}. Ying et al. \cite{ying2021image} uses robust data hiding to construct the immunized images. Users are able to locate tampers and conduct self-recovery.

In the real-world applications, the users wish to apply HI3 and expect high-quality image recovery after lossy data transmission. Thus, it requires an effective method able to simultaneously embed a same-sized image while ensuring artifact imperceptibility of the synthesized image as well as strong robustness. Lately, ADH-GAN \cite{yu2020attention} is the first to ensure robustness and large capacity at the same time. However, we find that the robustness is achieved by a hybrid method of both generating an attacked training dataset and directly adding noises on the original images. Therefore, data hiding of the secret image is performed after the attacks. In real-world application, the attacks are always performed after the data embedding. As a result, the robustness of \cite{yu2020attention} cannot be achieved in real-world applications.

Motivated by the short-comings of the previous works, we propose an end-to-end generative method, which robustly hides secret images into a cover image. 
We jointly train an embedding network, a decoupling network, and a revealing network. A double branched U-Net~\cite{ronneberger2015u} architecture is designed to implement the networks. We embed the secret images into the cover image. We use an improved real-world attack simulator to introduce adversarial training. On the recipient's side, we designed a decoupling network to extract the embedded residual image from the attacked image as well as recover the original cover image. Finally, the secret image can be recovered with high quality. Experimental results demonstrate that the proposed method can achieve high visual quality of both the marked and the recovered image despite the presence of different attacks. While ADH-GAN cannot hide multiple images under simple extension, we further embed two or three secret images into a cover image. The maximum payload of the proposed scheme is around 72bpp.
\begin{figure*}[!t]
\begin{center}
\includegraphics[width=1.0\textwidth]{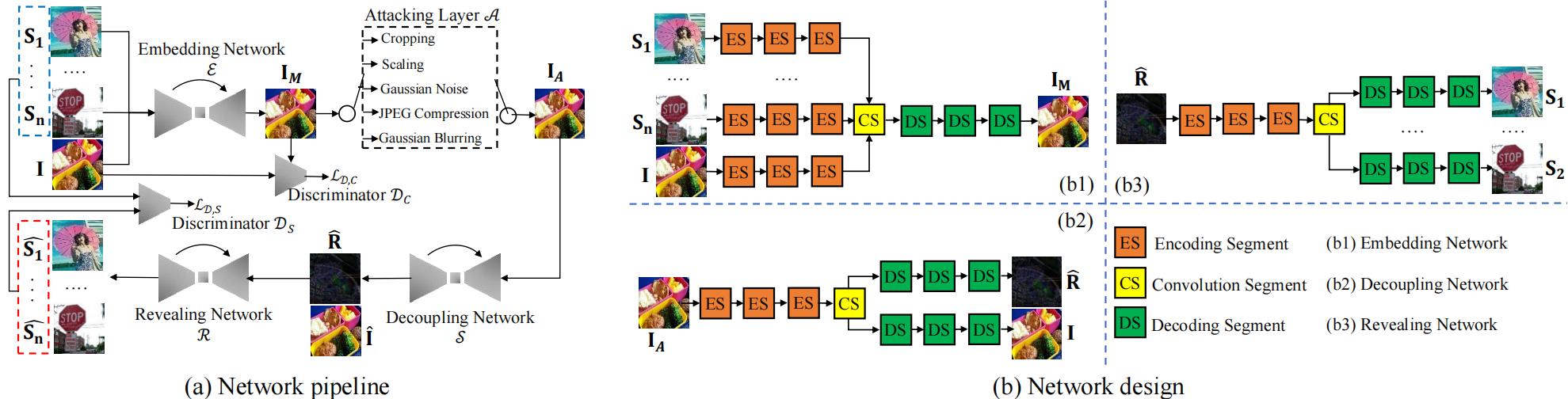}
\caption{Pipeline design. The embedding network hides the secret image into the cover image. The attacking layer simulates typical digital attacks. The image extraction is two-staged: we first extract the embedded residual, and then decode it into the secret image.}
\label{framework}
\end{center}
\end{figure*}

The main contributions of this paper are three-folded: 1) We are the first to robustly embed up to three secret images. 2) The proposed progressive image recovery method and the real-world attack simulation are effective in promoting the resilience against various digital attacks. The robustness of our scheme is superior compared the state-of-the-art scheme.3) Better qualities of the marked image and the recovered image can be achieved simultaneously.
\section{Proposed Method}
\label{section:method}
Fig.~\ref{framework} depicts the sketch of our proposed end-to-end deep network based HI3 method. The pipeline consists of an embedding network $\mathcal{E}$, an attacking layer $\mathcal{A}$, a decoupling network $\mathcal{S}$, a revealing network $\mathcal{R}$ and two discriminators $\mathcal{D}_{C}$ and $\mathcal{D}_{S}$. We propose a novel multi-branched U-Net architecture to implement $\mathcal{E}$, $\mathcal{S}$ and $\mathcal{R}$. The discriminators are to respectively monitor the quality of the marked images and the extracted images. We use Patch-GAN \cite{isola2017image} as the discriminators.
\subsection{Image embedding pipeline}
We use the embedding network $\mathcal{E}$ to hide one or more secret images $\mathbf{S}=\{\mathbf{S}_1,...,\mathbf{S}_n\}$ into a cover image $\mathbf{I}$. $n$ denotes the number of secret images and we let $n\in[1,3]$. A residual image $\mathbf{R}$ is generated by $\mathcal{E}$ and the marked image $\mathbf{I}_{M}$ is produced by adding the residual image on the cover image, i.e., $\mathbf{I}_{M}$=$\mathbf{I}$+$\mathbf{R}$. The residual image can be viewed as an compressed version of $\mathbf{S}$.
We extend the typical U-Net architecture by introducing multiple ($n$) independent encoding parts that respectively generate features for the cover image $\mathbf{I}$ and the secret image $\mathbf{S}$. These branches do not share weights. Fig.~\ref{framework}(b1) shows an example of $\mathcal{E}$ where there are two secret images and therefore three encoding branches. 
In the decoding part, the features from the branches are all concatenated with the output of the previous layer. The main advantage of this design compared to the original U-Net structure is that the relevance of the features from the secret images and the cover image can be disentangled and more efficiently utilized in generating an indistinguishable marked image with improved robustness.
\subsection{Improved Attack Simulation for Real-World Robustness}
We build the attacking layer $\mathcal{A}$ that attacks the marked image $\mathbf{I}_M$ as a way of adversarial training. Similar to \cite{zhu2018hidden,mun2019finding}, the common digital attacks are simulated by differentiable methods. The attacking layer contains five types of typical digital attacks, namely, the addition of Gaussian noise, the Gaussian blurring, the random scaling, the lossy JPEG compression and the random cropping. We take the implementation from \cite{zhu2018hidden} except: 1) we build our own JPEG simulator. 2) JPEG compression attack is always present, i.e., the other attacks are further conducted on the JPEG compressed image.

Though many schemes have already proposed a variety of JPEG simulators, e.g., JPEG-SS \cite{liu2021jpeg}, JPEG-Mask \cite{zhu2018hidden}, MBRS \cite{jia2021mbrs}, it is reported that the real-world JPEG robustness is still far from satisfactory. We analyze that the reason is mainly two-folded: 1) the networks will finally over-fit a fixed JPEG simulator, resulting in poor performance in blind JPEG artefact removal. 2) other attacks such as rescaling maintain way more clues ( $\mathbf{R}$) compared to JPEG compression. As a result, when the attacks are iteratively performed, the networks will sacrifice robustness against JPEG for the rest. 

To address the first issue, we propose to conduct a similar instance-agnostic JPEG interpolation, inspired by Mix-up \cite{zhang2017mixup} which uses label interpolation for data augmentation in classification tasks. We construct the simulated JPEG image using different implementations and quality factors as follows:
\begin{equation}
\label{equation_JPG}
\mathbf{I}_{jpg}=\sum_{\mathcal{J}_{k}\in \mathcal{J}}\sum_{QF_{l}\in[10,100]}\epsilon\cdot\mathcal{J}_{k}(\mathbf{I}_{M}, QF_{l}),
\end{equation}
where $\sum_{\mathcal{J}_{k}\in \mathcal{J}}\epsilon=1$, $\mathcal{J} \in $\{JPEG-SS, JPEG-Mask, MBRS\} and $QF$ stands for the quality factor. To address the second issue, when we train the network to be robust against the attacks other than JPEG compression, we performed the attacks on $\mathbf{I}_{jpg}$ instead of directly on $\mathbf{I}_{M}$, where we fix all of the $QF_{l}$ in (\ref{equation_JPG}) to be 90.
\subsection{Progressive image recovery pipeline}
Besides the improved JPEG simulator, we propose to progressively recover the secret by extracting the embedded residual $\mathbf{R}$ first. Successful image recovery from the attacked image has long been a difficult task to achieve for many previous schemes using straight-forward image recovery \cite{baluja2017hiding,baluja2019hiding,duan2019reversible}. 
From the observation that the information of the hidden secret images $\mathbf{S}$ within the cover image $\mathbf{I}$ can only be reconstructed by the residual image $\mathbf{R}$, we begin the progressive image recovery by first extract as well as augment the embedded residual image as $\hat{\mathbf{R}}$.

\begin{figure*}[!t]
	\centering
	\includegraphics[width=1.0\textwidth]{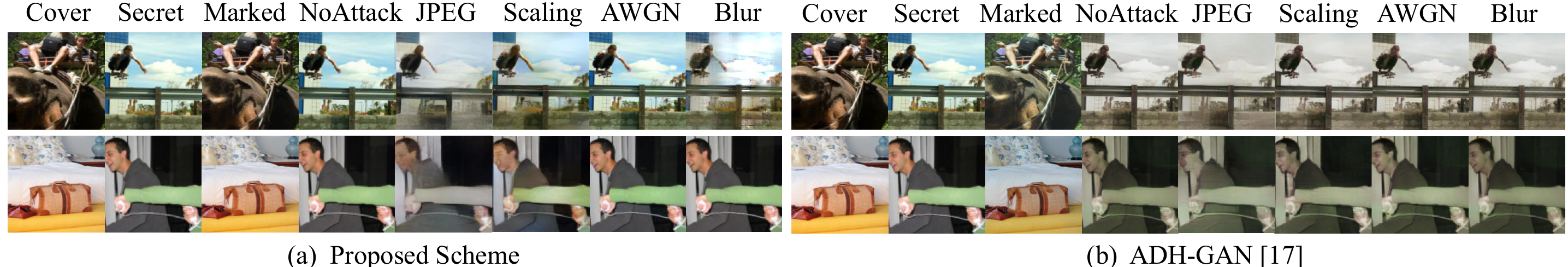}
	\caption{Comparison of robustness against various kinds of attacks.}
	\label{comparison}
\end{figure*}
\begin{figure}[!t]
	\centering
	\includegraphics[width=0.49\textwidth]{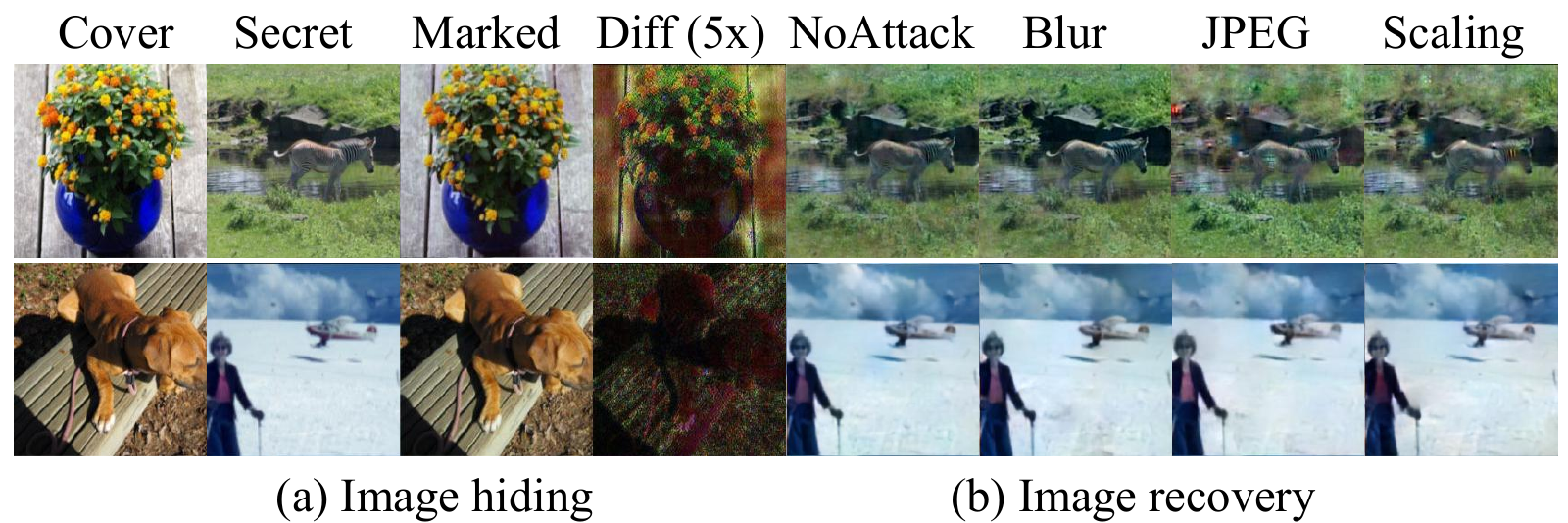}
	\caption{Image embedding and extraction test of hiding one secret image. The difference is augmented for better illustration.}
	\label{marked}
\end{figure}

We train a decoupling network $\mathcal{S}$ to extract $\hat{\mathbf{R}}$ ahead of the recovery of the secret images $\mathbf{S}$. During training stage, the decoupling network $\mathcal{S}$ has access to both the original embedded residual image $\mathbf{R}$ and the attacked image $\mathbf{I}_{A}$. Therefore, the network knows the difference between them and is able to learn a function to approximately transform $\mathbf{I}_{A}$ into $\hat{\mathbf{R}}$. Considering that the residual image $\mathbf{R}$ is usually weak, we encourage $\hat{\mathbf{R}}$ to be as close as possible as $5\times\mathbf{R}$. We believe the augmentation can help improving the learning performance. We also demand $\mathcal{S}$ to recover the original image to aid residual extraction. 
\begin{table}[!t]
	\caption{Performance comparison of image extraction over 1000 test images. (First row: PSNR. Second row:SSIM)}
	\label{table_crop}
	\begin{center}
	\begin{tabular}{c|c|c|c|c|c}
	\hline
		Method & NoAttack & JPEG & Scaling & Blur & C\&W\\
		\hline
		\multirow{2}{*}{Proposed} 
		& 29.76 & 26.69 & 28.47 & 28.32 & 26.73\\
		& 0.916 & 0.875 & 0.895 & 0.905 & 0.863\\
		\hline
		ADH-GAN & 25.22 & 17.64  & 17.41 & 14.63 & 16.58\\
		\cite{yu2020attention} & 0.867 & 0.634 & 0.625 & 0.498 & 0.588 \\
		\hline
	\end{tabular}
	\end{center}
\end{table}

Afterwards, we feed the revealing network $\mathcal{R}$ with $\hat{\mathbf{R}}$ and recovers the secret image $\hat{\mathbf{S}}$. $\hat{\mathbf{S}}=\{\hat{\mathbf{S}_1},...,\hat{\mathbf{S}_n}\}$. We also build $\mathcal{S}$ and $\mathcal{R}$ on top of the proposed multi-branch U-Net architecture. Fig.~\ref{framework}(b2,b3) shows an example. The difference compared to $\mathcal{E}$ is that we build $n$ branches for $\mathcal{R}$ and two branches for $\mathcal{S}$ in the decoding parts, while for each network there is only one shared encoding part.

\subsection{Objective loss function}

We encourage the marked image $\mathbf{I}_M$ and the extracted image $\mathbf{I}_R$ to respectively resemble the targeted image $\mathbf{I}$ and the secret image $\mathbf{S}$. Besides, we encourage $\mathcal{S}$ to approximately reconstruct $\mathbf{I}$. Therefore, the reconstruction loss is  $\mathcal{L}_{rec}=\mathcal{F}(\mathbf{I}, \mathbf{I}_M)+\mathcal{F}(\mathbf{S}, \hat{\mathbf{S}})+\mathcal{F}(\mathbf{I}, \hat{\mathbf{I}})$, where $\mathcal{F}$ is the $\ell_2$ distance. The adversarial loss is to further control the introduced distortion by fooling the discriminator. For the discriminative loss $\mathcal{L}_{dis,C}$ and $\mathcal{L}_{dis,R}$, we accept the least squared adversarial loss (LS-GAN) \cite{mao2017least}. Note that we train a unified $\mathcal{D}_{R}$ to distinguish $\mathbf{S}$ from $\hat{\mathbf{S}}$. Finally, the reconstruction loss for the extracted residual image is $\mathcal{L}_\mathcal{S}=\|5 \cdot \mathbf{R}-\hat{\mathbf{R}}\|_2^2$.
The total loss for the proposed scheme is $
\mathcal{L}_{\mathcal{G}}=\mathcal{L}_{rec}+\alpha\cdot\mathcal{L}_{dis,C}+\beta\cdot\mathcal{L}_{dis,R}+\gamma\cdot\mathcal{L}_\mathcal{S}$, where $\alpha$,$\beta$ and $\gamma$ are hyper-parameters.

\begin{figure*}[!t]
	\centering
	\includegraphics[width=1.0\textwidth]{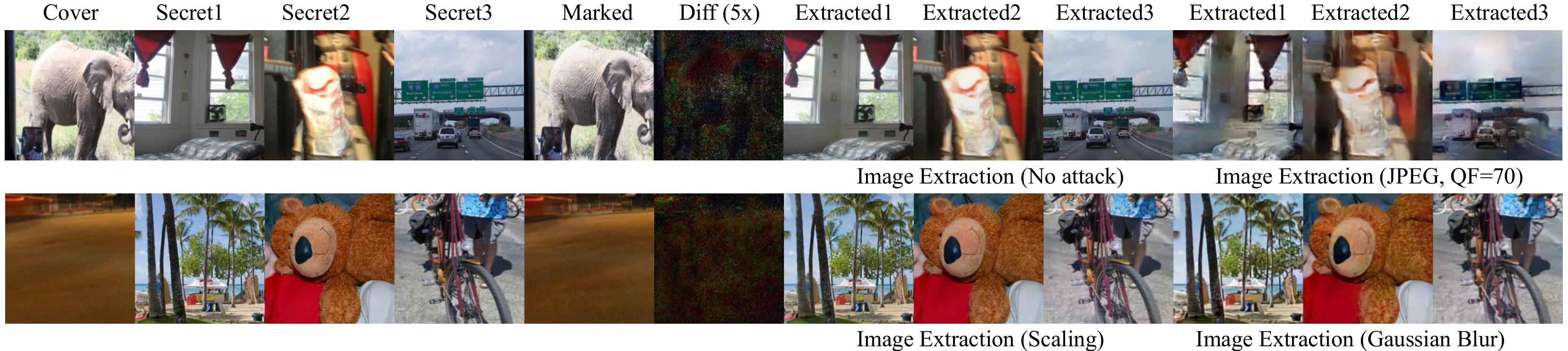}
	\caption{Comparison of robustness against different attack among different methods. ``GT'' stands for the ground-truth image.}
	\label{multiple_images}
\end{figure*}
\section{Experiments}
\label{section:experiment}

We train the scheme on the COCO training/test set \cite{lin2014microsoft}
We resize the images to the size of $256\times~256$. The hyper-parameters are set as $\alpha=0.05, \beta=0.05, \gamma=1.5$. The batch size is set as four. We find a larger batch size will jeopardize the networks from generalizing the learned robustness. The payload of the proposed scheme depends on the number of the hidden secret images, i.e., 48bpp (bit per pixel) when $n=2$ and 72bpp when $n=3$. We use Adam optimizer \cite{kingma2014adam} with the default parameters. The learning rate is $2\times10^{-4}$. The embedding rate of the state-of-the-art robust HI3 scheme (ADH-GAN~\cite{yu2020attention}) is 24bpp. Therefore, we begin our evaluation with hiding one image ($n=1$). Then we discuss on hiding multiple images ($n>1$). While ADH-GAN cannot hide multiple images under simple extension, we further embed two or three secret images into a cover image using our method. 

\subsection{Robustly Hiding One Image}
\noindent\textbf{Imperceptibility and Robustness. }
In Fig.~\ref{comparison} and Fig.~\ref{marked}, we arbitrarily select four pairs of images for illustration. The differences are augmented five times for better illustration. We can observe that the embedding of the secret image on the marked images is imperceptible to human eyes. The difference $\mathbf{D}$ is imperceptible. Little detail of the watermark can be found. We have conducted more embedding experiments over 1000 images and the average PSNR between the marked images and the cover images is 33.94dB. The average SSIM \cite{wang2004image} is 0.9517. In Fig.~\ref{marked}, we further apply different attacks on the marked images. The results are promising. We can clearly observe that the overall image quality of the extracted images is high, which does not downgrade much despite the presence of the attacks. The reason why the secret image cannot be losslessly extracted even when there is not attack is because of the adversarial training. The networks are frequently exposed to attacks and therefore they jointly learn a robust but not accurate way of image reconstruction. One can train his one networks without using the attacking layer in an attack-free environment.

\noindent\textbf{Comparison with ADH-GAN. }
In Fig.~\ref{comparison}, we compare our method with \cite{yu2020attention} on both robustness and imperceptibility. 
First, there exists color inconsistency in the marked image produced by \cite{yu2020attention}, which is also a problem for many previous works \cite{baluja2017hiding,zhu2018hidden}. However in our scheme, this phenomenon rarely occur.
We analyze that for successful data extraction, the data hiding networks sometimes weaken the strength of the targeted image apart from strengthening the embedded signal. Such occurrence can be largely lowered by penalizing a larger image reconstruction loss. However, the data extraction performance will be much worse.
Second, we can observe that in the extracted images of \cite{yu2020attention}, there is visual artifact both in color and in texture. In contrast, our scheme can largely preserve image fidelity without hallucinating image details. For space limit, please zoom in for a closer scrutiny.
We proceed our comparison by randomly selecting 1000 pairs of images and apply various kinds of attacks. For JPEG compression, we take $QF=70$ as an example. The average performances of the proposed scheme and \cite{yu2020attention} are reported in Table \ref{table_crop}. Note that unlike \cite{zhu2018hidden}, the performances are from one model trained with multiple kinds of attacks instead of from separate specified models. The average SSIMs of the recovered image produced by the proposed method are generally above 0.8 despite the variety of added attacks. C\&W \cite{carlini2017towards} is a famous adversarial example (AE) algorithm that the networks are not trained against. The robustness can be generalized against adversarial attacks.
By comparison, the average performance of \cite{yu2020attention} is worse than that of the proposed method. The result is consistent with Fig.~\ref{comparison} which proves that the proposed method has better imperceptibility and robustness in image hiding.

\begin{figure}[!t]
	\centering
	\includegraphics[width=0.49\textwidth]{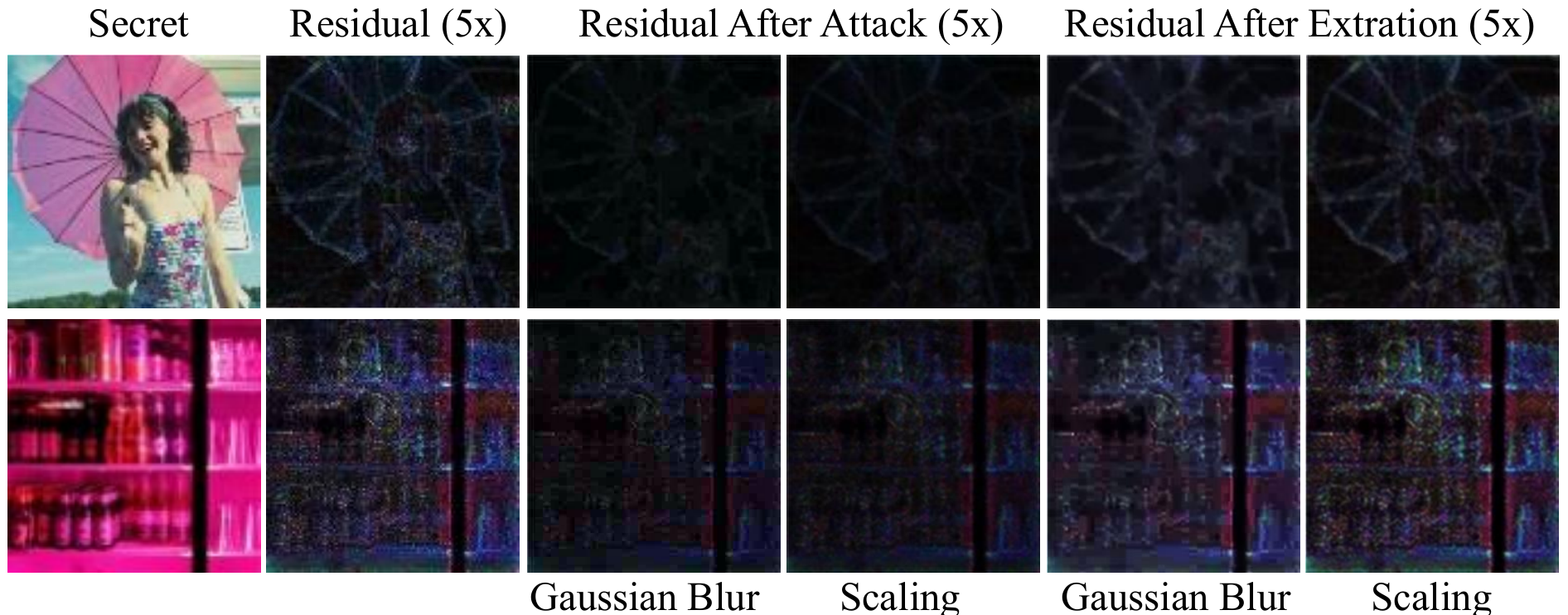}
	\caption{Examples of the extracted residual images. The embedded residual images are largely destroyed by the attacks. The decoupling network successfully augments them.}
	\label{afterstand}
\end{figure}
\begin{table}[!t]
	\renewcommand{\arraystretch}{1.3}
	\caption{Ablation study of image extraction with JPEG QF=70.}
	\label{table_ablation}
	\centering
	\begin{tabular}{c|cc}
		\hline
		Configuration & PSNR & SSIM \\
		\hline
        Without progressive recovery & 13.66  & 0.612 \\
        \hline
         Without any discriminator & 15.37  & 0.726\\
        \hline
        Implementing $\mathcal{E}$ using traditional U-Net & 18.16  & 0.817 \\
        \hline
         Without the proposed JPEG simulator & 23.95  & 0.839\\
         \hline
        Fully implemented & 26.68  & 0.875 \\
		\hline
	\end{tabular}
\end{table}
\begin{table}[!t]
	\caption{Performance of image extraction on hiding multiple images. The payload varies from 24bpp ($n=1$) to 72bpp ($n=3$)}
	\label{table_multiple_images}
	\begin{center}
	\begin{tabular}{c|c|c|c|c|c|c}
	\hline
		$n$ & NoAttack & JPEG & Scaling & Blur & AWGN & C\&W\\
		\hline
		\multirow{2}{*}{2} 
		& 28.62 & 24.60 & 26.58 & 26.82 & 25.01 & 26.71\\
		& 0.889 & 0.824 & 0.873 & 0.854 & 0.793 & 0.847 \\
		\hline
		\multirow{2}{*}{3}
		& 28.14 & 23.02  & 24.59 & 24.22 & 22.78 & 24.13\\
		& 0.862 & 0.757 & 0.854 & 0.844 & 0.735 & 0.806 \\
			\hline
		\multirow{2}{*}{4}
		& 23.79 & 17.81 & 19.17 & 16.00 & 13.64 & 18.93\\
		& 0.805 & 0.664 & 0.737 & 0.547 & 0.511 & 0.677\\
		\hline
	\end{tabular}
	\end{center}
\end{table}

\noindent\textbf{Validation of the Pipeline Design. }
We begin with analyzing what the decoupling network $\mathcal{S}$ can learn. Fig. \ref{afterstand} shows two examples where the embedding network $\mathcal{E}$ encodes the secret images (first column) into the residual images (second column) and adds them onto the cover images (not shown). In the third and fourth column, we show how the traditional attacks destroy the embedded residual image. However, after running $\mathcal{S}$, we can clearly observe that the extracted residual images are much close to the original residual images. Through the comparison, we prove that the decoupling network can effectively reconstruct the added residual images, which are crucial in recovering the secret images.

Then, we verify the effectiveness of several components in the scheme. In Table.~\ref{table_ablation}, we show the experimental results of several ablation tests. In each test, we train the modified pipeline using the same learning rates and optimizers. We stop the training until the reconstruction loss of the marked image in each test is close to that of the fully implemented proposed method, and the recovery loss cannot be further minimized. The results show that every component plays an important role.

\subsection{Robustly Hiding Two or Three Images}
Fig.~\ref{multiple_images} shows three groups of results on hiding three secret images. We see that the distortion introduced on the cover image is significantly and understandably larger than that on hiding a single image. Also, the residual images are no longer easily interpretable. But overall, the image quality of the marked image and the extracted images are high. Besides, we can barely observe one extracted image from another, indicating the extracted images are well disentangled.

Table.~\ref{table_multiple_images} shows the average results of 1000 groups of images on hiding two or three images. We retrain the networks with different amount of secret images. The average PSNR of the marked image is 32.13dB when $n=2$ and 32.69dB when $n=2$. We see that the image quality of the extracted images can be preserved. We also find that the performance does not drop much when we hide three images compared to Table.\ref{table_crop}. However, the performance significantly drops when we tend to hide a fourth secret image. Therefore, we believe the maximum payload of the proposed scheme is around 72bpp.

\section{Conclusion}
\label{section:conclusion}
This paper presents a new method for hiding images into images with real-world robustness. 
We propose an embedding network to conceal secret images into a cover image, where the introduced perturbation is close to imperceptible. We propose an improved attack simulator and progressively recover the hidden images. The comprehensive experiments prove that compared to the state-of-the-art methods, the proposed method can simultaneously embed a larger payload, ensure the visual effect of the marked image and offer a much stronger robustness.
\bibliographystyle{IEEEbib}
\bibliography{reference}

\end{document}